\documentstyle{llncs}
\begin{document}
\title{Prediction with Expert Advice in Games
with Unbounded One-Step Gains}

\author{Vladimir V. V'yugin}

\institute{Institute for Information Transmission Problems,
Russian Academy of Sciences,
Bol'shoi Karetnyi per. 19, Moscow GSP-4, 127994, Russia\\
e-mail vyugin@iitp.ru}

\maketitle
\begin{abstract}
The games of prediction with expert advice are considered in this paper.
We present some modification of Kalai and Vempala algorithm of
following the perturbed leader for the case of unrestrictedly large
one-step gains. We show that in general case the cumulative gain of
any probabilistic prediction algorithm can be much worse than the gain
of some expert of the pool.
Nevertheless, we give the lower bound for this cumulative gain in
general case and construct a universal algorithm which
has the optimal performance; we also prove that
in case when one-step gains of experts of the pool
have ``limited deviations'' the performance of our algorithm is close to
the performance of the best expert.
\end{abstract}

\section{Introduction}\label{intr-1}

Experts algorithms are used for online prediction or repeated decision making
or repeated game playing. Any such algorithm is based on a ``pool of
experts''. At any step $t$, each expert gives its recommendation. From this,
a ``master decision'' is performed. After that, losses (or rewards) $s_t^i$
are assigned to each expert $i=1,\dots ,m$ by the environment (or adversary).
The master algorithm also receives some loss or reward depending on
the master decision. The goal of the
master algorithm is to perform almost as well as the best expert
in hindsight in the long run.

Prediction with Expert Advice considered in this paper proceeds as follows.
We are asked to perform
sequential actions at times $t=1,2,\dots, T$. At each time step $t$, we
observe results of actions of experts in the form of their gains and losses
on steps $<t$. After that, at the beginning
of the step $t$ {\it Learner} makes a decision to follow one of these experts,
say Expert $i$. At the end of step $t$ Learner receives
the same gain or loss as Expert $i$ at step $t$.

We use notations and definitions from~\cite{HuP2004} and~\cite{KaV2003}.
Let $s^i_{1:t}=s^i_1+\dots +s^i_t$ be the cumulative loss of Expert $i$
at time $t$.
Given $s^i_{1:t-1}$, $i=1,\dots ,m$, at time $t$, a natural idea
to solve the expert problem is
``to follow the leader'', i.e. to select the expert $i$ which performed
best in the past.
The following simple example from Kalai and Vempala~\cite{KaV2003}
shows that Learner can perform much worse than each expert:
let the current losses of two experts on steps $t=1,\dots ,6$ be
$s^1_{1,2,3,4,5,6}=(0,1,0,1,0,1)$ and
$s^2_{1,2,3,4,5,6}=(\frac{1}{2},0,1,0,1,0)$. The ``Follow Leader''
algorithm always chooses the wrong prediction.

The method of following the perturbed leader was discovered by
Hannan~\cite{Han57}. Kalai and Vempala~\cite{KaV2003} rediscovered
this method and published a simple proof
of the main result of Hannan. They called the algorithm of this type
FPL (Following the Perturbed Leader).
Hutter and Poland~\cite{HuP2004} presented a further developments of
the FPL algorithm for countable class of experts, arbitrary weights
and adaptive learning rate.

The FPL algorithm outputs prediction of an expert $i$ which minimizes
$$
s^i_{1:t-1}-\frac{1}{\epsilon}\xi_t^i,
$$
where $\xi_t^i$, $i=1,\dots m$, $t=1,2,\dots$,
is a sequence of i.i.d random variables distributed according
to the exponential distribution with the density $p(t)=e^{-t}$,
and $\epsilon$ is {\it a learning rate}. Kalai and Vempala~\cite{KaV2003}
show that the expected cumulative loss of the FPL algorithm has the upper bound
$$
E(s_{1:t})\le (1+\epsilon)\min\limits_{i=1,\dots , n}
s^i_{1:t}+\frac{\log n}{\epsilon},
$$
where $\epsilon$ is a learning rate, $n$ is the number of experts..

In the papers cited above the loss of each expert $i$
can change at any step $t$ by a bounded quantity, for example,
$0\le s^i_t\le 1$ for all $t$.
Poland and Hutter~\cite{PoHP2005} extended this analysis for games with
one-step losses upper bounded by an increasing sequence $B_t$ given in advance, i.e.,
$s_t\le B_t$ for all $t$. Allenberg et al.~\cite{AAAGO2006} also considered unbounded
losses, but with different algorithm than in this paper.

In games considered in this paper the players will incur gains (loss is
a negative gain); $s^i_t$ denotes one-step gain of a player $i$.
For practical purposes, the property $0\le s^i_t\le 1$ seems to be
too restrictive.

In Appendix~\ref{brown-1} we consider some applications of results of
Sections~\ref{positive-loss}-\ref{negative-loss} of this paper.
We define two financial experts learning the fractional Brownian motion whose
one-step gains at any step can not be restricted in advance.
This application is at the bottom of our special interest in
zero-sum games with unbounded gains in Section~\ref{negative-loss}.

In this paper we present some modification of Kalai and Vempala
algorithm for the case of unrestrictedly large one-step gains not
bounded in advance.
We show that in general case, the cumulative gain of any
probabilistic prediction algorithm can be much worse than the gain
of some expert of the pool. Nevertheless, we give the lower bound
for cumulative gain of any probabilistic algorithm in general case
and prove that our universal algorithm has optimal performance; we
also prove that in case when one-step gains of experts of the pool
have ``limited deviations'' (in particular, when they are bounded)
the performance of our algorithm is close to the performance of
the best expert. This result is some improvement of results
mentioned above.

\section{Learning in games of two experts with unbounded gains}
\label{positive-loss}

In this section we give some preliminary results presenting the bounds
on the performance of the algorithm constructed in Section~\ref{sect-3}.

We consider a simple game $G$ of prediction with expert advice
by following of two experts with unbounded one-step gains.
The goal of the master algorithm
is to receive a cumulative gain not much worse than
the gain of the best expert in hindsight.

At each step $t$ of the game both experts receive the nonnegative one-step
gains $s^1_t$ and $s^2_t$, and their cumulative gains after step $t$ are equal to
$s^1_{1:t}=s^1_{1:t-1}+s^1_t$ and $s^2_{1:t}=s^2_{1:t-1}+s^2_t$.

For simplicity, we consider a variant when at each step $t$ of the game $G$
only one expert can receive a nonnegative one-step gain $s_t$, and the total
gain of the other expert is unchanged, i.e.,
$s^1_{1:t}=s^1_{1:t-1}+s_t$ and $s^2_{1:t}=s^2_{1:t-1}$ or
$s^2_{1:t}=s^2_{1:t-1}+s_t$ and $s^1_{1:t}=s^1_{1:t-1}$.
In the general case the analysis is similar.

We also consider {\it non-degenerate} experts (games), i.e., such that\\
$\max\{s^1_{1:t},s^2_{1:t}\}\to\infty$ as $t\to\infty$.

A probabilistic algorithm of following the leader
in the game with two experts is based on a computable function $f$ which
given cumulative gains $s^1_{1:t-1}$ and $s^2_{1:t-1}$ of the experts
in hindside outputs the probability of following the first expert
$P\{I=1\}=f(s^1_{1:t-1},s^2_{1:t-1})$ and the probability of following
the second expert $P\{I=2\}=1-P\{I=1\}$.

The analysis in case when these probabilities depend of the whole history
of gains is similar.

Let two experts be given. The {\it master} algorithm works as follows.

{\bf Probabilistic algorithm of following the leader}.

\noindent FOR $t=1,\dots T$

Given past cumulative gains of the experts
$s^1_{1:t-1}$ and $s^2_{1:t-1}$ choose
the expert $i\in\{1,2\}$ with probability $P\{I=i\}$.

Receive the one-step gains at step $t$ of two experts $s^1_t$ and $s^2_t$
and define one step gain $s_t=s^i_t$ of the master algorithm.

\noindent ENDFOR

The following theorem says that if a probabilistic algorithm of
following the leader has high performance in games with
bounded one-step gains then its performance in games with unbounded
one-step gains can be much worse than the performance of some experts.

\begin{theorem}\label{perf-1}
Let $\delta,\delta'$ be arbitrary close and arbitrary small
positive real numbers such that $\delta'>\delta$, and let
for any two non-degenerate experts with bounded one-step gains $s^i_t$, $i=1,2$,
i.e., such that $0\le s^i_t\le 1$ for all $t$,
a master algorithm has the expected cumulative gain
\begin{eqnarray}\label{perf-2}
E(s_{1:t})\ge (1-\delta)\max\limits_{i=1,2}s^i_{1:t}
\end{eqnarray}
for all sufficiently large $t$.
Then there exist two experts with unbounded one-step gains such that
the expected cumulative gain of the master algorithm is bounded from above
\begin{eqnarray}\label{perf-3}
E(s_{1:t})\le\delta'\max\limits_{i=1,2}s^i_{1:t}
\end{eqnarray}
for infinitely many $t$.
\end{theorem}
{\it Proof}. Let a master algorithm be given, and let
$P\{I=1\}=f(s^1,s^2)$ and $P\{I=2\}=1-P\{I=1\}$ be probabilities
to choose the best expert from two experts with cumulative gains $s^1$, $s^2$.
The proof of the theorem uses the following lemma.

\begin{lemma}\label{l-1}
Let $\delta,\delta'$ be positive real numbers such that
$\delta'>\delta$ and for any two experts with bounded one-step gains
the master algorithm has the expected performance (\ref{perf-2})
for all sufficiently large $t$.
Then for any two real numbers $\tilde s^1$ and $\tilde s^2$ a number $s^1$
exists such that $s^1\ge \tilde s^1$, $s^1\ge \tilde s^2$, and
$P\{I=1\}\ge 1-\delta'$, where
$P\{I=1\}=f(s^1,\tilde s^2)$ (and $P\{I=2\}=1-P\{I=1\}$).
\end{lemma}
{\it Proof}. Suppose that for some pair $\tilde s^1,\tilde s^2$ of real numbers
the contrary statement holds.
Then we can construct two experts with cumulative gains
$s^1_{1:t-1},s^2_{1:t-1}$, $t=1,2,\dots$, and
with step-gains equal $0$ or $1$ such that (\ref{perf-2}) is violated.

Define the sequences
$s^1_t,s^2_t$, $t=1,2,\dots t_0$, such that $s^1_t,s^2_t$ are equal to $0$
or $1$ and such that $s^1_{1:t_0}=\tilde s^1$ and $s^2_{1:t_0}=\tilde s^2$
for some $t_0$.
After that, define $s^1_t=1$ and $s^2_t=0$ for all $t>t_0$.
We have $s^1_{1:t-1}>s^2_{1:t-1}$ and $P\{I=1\}<1-\delta'$
for all sufficiently large $t$. Then
for the expected one-step gain of the master algorithm,
$$
E(s_t)<1\circ P\{I=1\}+0\circ P\{I=2\}\le 1-\delta'=s^1_t(1-\delta')
$$
holds for all these $t$. Since $s^1_{1:t}\to\infty$ as $t\to\infty$,
we have $E(s_{1:t})<(1-\delta)s^1_{1:t}$ for all sufficiently
large $t$.
This is a contradiction with (\ref{perf-2}). Hence, for some $t$
we have $s^1_{1:t-1}>s^2_{1:t-1}$ and $P\{I=1\}\ge 1-\delta'$, where
$P\{I=1\}=f(s^1_{1:t-1},s^2_{1:t-1})$.
$\triangle$

We define two experts with unbounded one-step gains as follows.
Define $s^1_0=0$ and $s^2_0=0$. By Lemma~\ref{l-1} a number
$s^1>0$ exists such that $P\{I=1\}\ge 1-\delta$, where
$P\{I=1\}=f(s^1,0)$. Define $s^1_1=s^1$, $s^2_1=0$.

Let $t$ be even, and let $s^1_{1:t-1}$ and $s^2_{1:t-1}$ be defined
on previous steps. We will use the induction hypothesis:
$s^1_{1:t-1}>s^2_{1:t-1}$ and $P\{I=1\}\ge 1-\delta$.
By definition this induction hypothesis holds
$t=2$. Define one-step gains of experts 1 and 2 at step $t$:
$s^1_t=0$ and $s^2_t=M_t$, where
$M_t=\frac{E(s_{1:t-1})}{\delta'-\delta}$ and $E(s_{1:t-1})$ is the
mathematical expectation of the cumulative gain of the master algorithm
on steps $<t$.

Let $t$ be odd. By Lemma~\ref{l-1} a number $s^1$ exists such that
$s^1\ge s^1_{1:t-1}$, $s^1\ge s^2_{1:t-1}$, and $P\{I=1\}\ge 1-\delta$,
where $P\{I=1\}=f(s^1,s^2_{1:t-1})$.
Define $s^1_t=s^1-s^1_{1:t-1}$ and $s^2_t=0$. Then $s^1_{1:t}=s^1$ and
$s^2_{1:t}=s^2_{1:t-1}$. Evidently, the induction hypothesis is valid
after step $t$.

Let us prove that this definition is correct.
Let $t$ be even. By the induction hypothesis $s^1_{1:t-1}>s^2_{1:t-1}$ and
$P\{I=1\}\ge 1-\delta$, where $P\{I=1\}=f(s^1_{1:t-1},s^2_{1:t-1})$.
Then $P\{I=2\}<\delta$.
By definition $s^1_{1:t}=s^1_{1:t-1}$ and $s^2_{1:t}=s^2_{1:t-1}+M_t$.
Then we obtain an upper bound for the expected
one-step gain of the master algorithm
\begin{eqnarray*}
E(s_t)=s^1_t E\{I=1\}+s^2_t E\{I=2\}\le\delta M_t.
\end{eqnarray*}
For expected cumulative gain, we have
\begin{eqnarray}
E(s_{1:t})\le E(s_{1:t-1})+\delta M_t\le
\delta'(s^2_{1:t-1}+M_t)=\delta's^2_{1:t}.
\label{byd-1}
\end{eqnarray}
Inequality (\ref{byd-1}) holds for all even steps $t$.
$\triangle$

Decreasing the lower bound of the performance of a probabilistic algorithm
for games with
bounded one-step gain functions we can increase it for games with
unbounded gain functions. The limit case is given by the following
simple example.
Evidently, the expected cumulative gain of the probabilistic
algorithm which chooses one
of two experts with equal probabilities $\frac{1}{2}$ has the lower bound
\begin{eqnarray}\label{triv-1}
E(s_{1:t})=\frac{1}{2}s^1_{1:t}+\frac{1}{2}s^2_{1:t}\ge
\frac{1}{2}\max\limits_{i=1,2}s^i_{1:t}
\end{eqnarray}
for $i=1,2$.

The following simple diagonal argument shows that the cumulative gain of
any probabilistic algorithm of following the leader can be bigger than
this bound for some experts, analogously, it can be smaller for some experts.

\begin{proposition}\label{prop-1}
For any $\delta$ such that $0<\delta<1$ and for any probabilistic
algorithm of following the best expert, two experts exist such that
the expected cumulative gain of this algorithm satisfies
\begin{eqnarray}
E(s_{1:t})\le
\frac{1}{2}(1+\delta)\max\limits_{i=1,2}s^i_{1:t},
\label{fir-1}
\end{eqnarray}
for all sufficiently large $t$, where $s^1_{1:t}$, $s^2_{1:t}$ are
cumulative gains of these experts. Analogously, two experts exist
such that
\begin{eqnarray}
E(s_{1:t})\ge
\frac{1}{2}(1-\delta)\max\limits_{i=1,2}s^i_{1:t}
\label{fir-2}
\end{eqnarray}
for all sufficiently large $t$.
\end{proposition}
{\it Proof}.
Given a probabilistic algorithm of following the best expert and $\delta$
such that $0<\delta<1$ define recursively the gains of expert 1 and
expert 2 at any step $t$ as follows.
Let $s^1_{1:t-1}$ and
$s^2_{1:t-1}$ be cumulative gains of these experts incurred at steps $<t$.
Let $M_t=E(s_{1:t-1})/\delta$, where $E(s_{1:t-1})$ is the expected cumulative
gain of the master algorithm in the past.

If $P\{I=1\}>\frac{1}{2}$ then define $s^1_t=0$ and $s^2_t=M_t$, and define
$s^1_t=M_t$ and $s^2_t=0$ otherwise. Then
$E(s_t)=s^1_tP\{I=1\}+s^2_tP\{I=2\}\le\frac{1}{2}M_t$ and
$E(s_{1:t})=E(s^1_{1:t-1})+E(s_t)\le\frac{1}{2}(1+\delta)M_t\le
\frac{1}{2}(1+\delta)\max\limits_{i=1,2}s^i_{1:t}$ for all sufficiently
large $t$.

To prove (\ref{fir-2}) define $s^1_t=M_t$ and $s^2_t=0$ if
$P\{I=1\}>\frac{1}{2}$, and define $s^1_t=0$ and $s^2_t=M_t$ otherwise.
The following derivation is analogous to the proof of (\ref{fir-1}).
$\triangle$

\section{Asymptotically optimal algorithm of following the perturbed leader}
\label{sect-3}

In this section we show that the bounds (\ref{perf-2}) and  (\ref{perf-3})
obtained in Theorem~\ref{perf-1} can be achieved by some probabilistic
algorithm. More correctly, for any $\delta>0$ using the
method of following the perturbed leader
we construct a universal algorithm such that for any $\delta$
such that $0<\delta<1$ the lower bound
\begin{eqnarray*}
E(s_{1:t})\ge (1-\delta)\max\limits_{i=1,2}s^i_{1:t}
\end{eqnarray*}
is valid for all sufficiently large $t$ for arbitrary two experts
($i=1,2$) with bounded one-step gain functions (and even in more general case),
and, at the same time, for some $\delta'>0$
the bound
\begin{eqnarray*}
E(s_{1:t})\ge\delta'\max\limits_{i=1,2}s^i_{1:t}
\end{eqnarray*}
is valid for all experts with arbitrary unbounded one-step
gain functions. Here $E(s_{1:t})$ is the cumulative expected
gain of the master algorithm.

Note that in this section the cumulative gain is always nonnegative
$s^i_{1:t}\ge 0$
for all $t$ and for $i=1,2$. In Section~\ref{negative-loss} we consider
the case when the gains can be negative, i.e., experts can incur losses.
Recall that, for simplicity, we suppose that at any step $t$ only one expert
can receive a positive one-step gain, i.e., $s^1_t=0$ or $s^2_t=0$.
We denote $s_t=\max\{s^1_t,s^2_t\}$.

Let $\xi^1_1,\xi^2_1,\xi^1_2,\xi^2_2,\dots$ be a sequence of
i.i.d. random variables
distributed according to the exponential law with the density $p(t)=e^{-t}$.

We consider the FPL algorithm with learning rate
\begin{eqnarray}\label{eps-1}
\epsilon_{t-1}=\frac{1}{\mu\max\{s^1_{1:t-1},s^2_{1:t-1}\}},
\end{eqnarray}
where $t=1,2,\dots$ and $\mu$, where $0<\mu<1$, is a parameter of the
algorithm.

We suppose without loss of generality that $s^1_0=s^2_0=1$.
By definition the sequence $\epsilon_1,\epsilon_2,\dots$ is non-decreasing.

The FPL algorithm is defined as follows:

{\bf FPL algorithm}.

\noindent FOR $t=1,\dots T$

Output prediction of expert $i=i_{max}$ which maximizes
\begin{eqnarray}\label{m-1}
s^i_{1:t-1}+\frac{1}{\epsilon_{t-1}}\xi_t^i,
\end{eqnarray}
where $i=1,2$, $\epsilon_{t-1}$ is defined by (\ref{eps-1}).

Receive one-step gains $s_t^i$ for experts $i=1,2$,
and define one step gain $s^{i_{max}}_t$ of the master algorithm.

\noindent ENDFOR

Recall that a game $G$ of two experts is called non-degenerate if
$v_t=\max\{s^1_{1:t},s^2_{1:t}\}\to\infty$ as $t\to\infty$,
where $s^i_{1:t}$ is the cumulative gain of the expert $i=1,2$ at step $t$.
The number
\begin{eqnarray}\label{dev-1}
{\rm Dev}(G)=\limsup\limits_{t\to\infty}\frac{s_t}{v_t},
\end{eqnarray}
where $s_t=\max\{s^1_t,s^2_t\}$, is called {\it the deviation} of the game $G$.
For any game ${\rm Dev}(G)\le 1$ by definition.
In any non-degenerate game $G$ with bounded one-step gain function,
i.e. such that $0\le s_t\le A$ for all $t$ ($A$ is a positive real number),
${\rm Dev}(G)=0$.

\begin{theorem}\label{cor-2}
For any $\mu$ such that $0<\mu<1$ an FPL algorithm can be
specified such that for any non-degenerate game of two experts
its expected cumulative gain at any step $T$ has the lower bound
\begin{eqnarray}\label{thh-3}
l_{1:T}\ge e^{-\frac{2}{\mu}}(1-\mu)\max_{i=1,2}s^i_{1:T},
\end{eqnarray}
where $s^i_{1:T}$ is the cumulative gain of the expert $i=1,2$.

If ~${\rm Dev}(G)\le\frac{1}{2}\mu\delta$ for some $0<\delta<1$ then
\begin{eqnarray}
l_{1:T}\ge (1-\delta)(1-\mu)\max_{i=1,2}s^i_{1:T}
\label{thh-2}
\end{eqnarray}
holds for all sufficiently large $T$.
\footnote
{
The optimal value of $\mu$ for (\ref{thh-3}) is $\mu=0.618$. Then
$l_{1:T}\ge 0.015\max_{i=1,2}s^i_{1:T}$ in (\ref{thh-3}) and
$l_{1:T}\ge 0.382(1-\delta)\max_{i=1,2}s^i_{1:T}$ in (\ref{thh-2}).
Comparing these bounds with
(\ref{triv-1}), we reveal a large gap between bounds (\ref{thh-3})
and (\ref{thh-2}).
Author does not know if we can increase the lower bound
(\ref{thh-3}) when $\mu\approx\frac{1}{2}$ in (\ref{thh-2}).
}
\end{theorem}
{\it Proof}. This theorem will follow from Theorem~\ref{lemma-1}
and Corollary~\ref{cor-1} below. In the proof we follow the proof-scheme
of~\cite{HuP2004} and~\cite{KaV2003}.
$\triangle$

The analysis of optimality
of the FPL algorithm is based on an intermediate predictor IFPL
(Infeasible FPL) with the learning rate
\begin{eqnarray}\label{eps-2}
\epsilon_t=\frac{1}{\mu v_t},
\end{eqnarray}
where $v_t=\max\{s^1_{1:t},s^2_{1:t}\}$.

{\bf IFPL algorithm}.

\noindent FOR $t=1,\dots T$

Output prediction of expert $i=i_{max}$ with maximal value of
$$
s^i_{1:t}+\frac{1}{\epsilon_t}\xi_t^i,
$$
 where $i=1,2$,
$\epsilon_t$ is defined by (\ref{eps-2}), and
$\xi_t^1$, $\xi_t^2$ are independent random variables
distributed according to the exponential distribution with the
density $p(t)=e^{-t}$.

Receive one-step gains $s_t^i$ for experts $i=1,2$,
and define one step gain $s^{i_{max}}_t$ of the master algorithm.

\noindent ENDFOR

The IFPL algorithm predicts under the knowledge of $s^1_{1:t}$ and
$s^2_{1:t}$ ($\epsilon_t$ is their maximum), which both may not be
available at beginning of step $t$. Using unknown value of
$\epsilon_t$ (like $s^i_{1:t}$, $i=1,2$) is the main peculiarity
of our version of IFPL.

To distinguish the gains of the FPL and IFPL algorithms we denote
$s_t^I$ a one-step gain of the FPL algorithm at step $t$ and
$s_t^J$ is a one-step gain of the IFPL algorithm.
The expected one-step gains of the FPL and IFPL algorithms at the
step $t$ are denoted $l_t=E_t(s_t^I)$ and $r_t=E_t(s_t^J)$.

\begin{theorem}\label{lemma-1}
For any $\mu$, $0<\mu<1$, the expected one-step gain $l_t$ of the
FPL algorithm with learning rate (\ref{eps-1}) and the expected
one-step gain $r_t$ of the IFPL algorithm with learning rate
(\ref{eps-2}) satisfy the inequalities
\begin{eqnarray}
l_t\ge e^{-\frac{2}{\mu}}r_t
\label{thh-1}
\end{eqnarray}
for all $t$.

If ${\rm Dev}(G)\le\frac{1}{2}\mu\delta$ for some $0<\delta<1$ then
\begin{eqnarray}
l_t\ge (1-\delta)r_t
\label{thh-2a}
\end{eqnarray}
holds for all sufficiently large $t$.
\end{theorem}
{\it Proof}. For any $t>0$, denote $\xi^1=\xi^1_t$, $\xi^2=\xi^2_t$
and consider two random variables
\[
I=
  \left\{
    \begin{array}{l}
      1 \mbox{ if }
       s^1_{1:t-1}+\frac{1}{\epsilon_{t-1}}\xi^1>
       s^2_{1:t-1}+\frac{1}{\epsilon_{t-1}}\xi^2
    \\
      2 \mbox{ otherwise }
    \end{array}
  \right.
\]
and
\[
J=
  \left\{
    \begin{array}{l}
      1 \mbox{ if }
       s^1_{1:t}+\frac{1}{\epsilon_t}\xi^1>
       s^2_{1:t}+\frac{1}{\epsilon_t}\xi^2
    \\
      2 \mbox{ otherwise }
    \end{array}
  \right.
\]
Recall that $v_t=\max\{s^1_{1:t},s^2_{1:t}\}$ for all $t$.
For any real number $r$ we compare conditional probabilities
$P\{I=1|\xi^2=r\}$ with $P\{J=1|\xi^2=r\}$ and $P\{I=2|\xi^2=r\}$ with
$P\{J=2|\xi^2=r\}$.

In our analysis, the nontrivial cases are
$s^2_{1:t}=s^2_{1:t-1}+s_t$ and $s^1_{1:t}=s^1_{1:t-1}$
or $s^1_{1:t}=s^1_{1:t-1}+s_t$ and $s^2_{1:t}=s^2_{1:t-1}$, where $s_t>0$
(we indicate these cases in (\ref{ii-1})-(\ref{1-3}) below by $\pm$).
In this case the following chain of equalities is valid:
\begin{eqnarray}
P\{I=1|\xi^2=r\}=P\{s^1_{1:t-1}+\frac{1}{\epsilon_{t-1}}\xi^1>
s^2_{1:t-1}+\frac{1}{\epsilon_{t-1}}r|\xi^2=r\}=~~~~~~
\nonumber
\\
P\{\xi^1>\epsilon_{t-1}(s^2_{1:t-1}-s^1_{1:t-1})+r|\xi^2=r\}=~~~~~~
\nonumber
\\
P\{\xi^1>\epsilon_t(s^2_{1:t-1}-s^1_{1:t-1})+
(\epsilon_{t-1}-\epsilon_t)(s^2_{1:t-1}-s^1_{1:t-1})+r|\xi^2=r\}=~~~~~~
\label{1-1}
\\
e^{-(\epsilon_{t-1}-\epsilon_t)(s^2_{1:t-1}-s^1_{1:t-1})}
P\{\xi^1>\frac{1}{\mu v_t}(s^2_{1:t-1}-s^1_{1:t-1})+r|\xi^2=r\}=~~~~~~
\label{ii-1}
\\
e^{-\left(\frac{1}{\mu v_{t-1}}-\frac{1}{\mu v_t}\right)
(s^2_{1:t-1}-s^1_{1:t-1})\pm\frac{s_t}{\mu v_t}}~~~~~~
\nonumber
\\
P\{\xi^1>\frac{1}{\mu v_t}(s^2_{1:t-1}\pm s_t-s^1_{1:t-1})+r|\xi^2=r\}=~~~~~~
\label{1-2}
\\
e^{
\frac{s_t}{\mu v_t}
\left(\gamma_t\frac{s^1_{1:t-1}-s^2_{1:t-1}}{v_{t-1}}\pm 1\right)}
P\{\xi^1>\frac{1}{\mu v_t}(s^2_{1:t}-s^1_{1:t})+r|\xi^2=r\}=~~~~~~
\\
e^{\frac{s_t}{\mu v_t}
\left(\gamma_t\frac{s^1_{1:t-1}-s^2_{1:t-1}}{v_{t-1}}\pm 1\right)}
P\{J=1|\xi^2=r\}.~~~~~~
\label{1-3}
\end{eqnarray}
Here we have used twice, in (\ref{1-1})-(\ref{ii-1}) and in
(\ref{ii-1})-(\ref{1-2}), the equality $P\{\xi>a+b\}=e^{-b}P\{\xi>a\}$
for any random variable $\xi$ distributed according to the exponential law;
we also used the equality $v_t-v_{t-1}=\gamma_ts_t$, where
$0\le\gamma_t\le 1$, in (\ref{ii-1}).
The exponent (\ref{1-3}) is bounded
\begin{eqnarray}\label{ij-1}
2\ge\gamma_t\frac{s^1_{1:t-1}-s^2_{1:t-1}}
{v_{t-1}}\pm 1\ge -2.
\end{eqnarray}
These bounds follow from the inequalities $s^i_{1:t-1}/v_{t-1}\le 1$ and
$s^i_{1:t-1}\ge 0$ for all $t$ and for $i=1,2$.
We also used the inequality $s_t/v_t\le 1$ for all $t$.
Therefore,
\begin{eqnarray}
e^{\frac{2}{\mu}}P\{J=1|\xi^2=r\}\ge
P\{I=1|\xi^2=r\}\ge e^{-\frac{2}{\mu}}P\{J=1|\xi^2=r\}.
\label{bo-1}
\end{eqnarray}
Since, the the inequality (\ref{bo-1}) holds for all $r$,
it also holds unconditionally
\begin{eqnarray}
e^{\frac{2}{\mu}}P\{J=1\}\ge
P\{I=1\}\ge e^{-\frac{2}{\mu}}P\{J=1\}.
\label{i-00}
\end{eqnarray}
Analogously, we obtain
\begin{eqnarray}
e^{\frac{2}{\mu}}P\{J=2\}\ge P\{I=2\}\ge e^{-\frac{2}{\mu}}P\{J=2\}
\label{i=0}
\end{eqnarray}
for all $t=1,2,\dots$.

If ${\rm Dev}(G)\le\frac{1}{2}\mu\delta$,
when for sufficiently large $t$ the exponent (\ref{1-3})
is bounded from below by
$$
e^{-\frac{2}{\mu}\frac{s_t}{v_t}}\ge e^{-\delta}\ge 1-\delta.
$$
From this (\ref{thh-2a}) follows.

From (\ref{i-00}) and (\ref{i=0}) we obtain the lower bound (\ref{thh-1})
\begin{eqnarray}
l_t=E(s^I_t)=s^1_tP(I=1)+s^2_t P(I=2)\ge
\nonumber
\\
s^1_t e^{-\frac{2}{\mu}}P(J=1)+s^2_t e^{-\frac{2}{\mu}}P(J=2)=
e^{-\frac{2}{\mu}}E(s^J)=
e^{-\frac{2}{\mu}}r_t.
\label{expect-1}
\end{eqnarray}
$\triangle$



The connection between expected cumulative gain of the IFPL algorithm
$$
r_{1:T}=\sum\limits_{t=1}^T r_t
$$
and expected cumulative gain of the FPL algorithm
$$
l_{1:T}=\sum\limits_{t=1}^T l_t.
$$
is given in the following corollary.
\begin{corollary}\label{ifpl-fpl}
For any $\mu$ and $\eta$, $0<\mu,\eta<1$, the expected cumulative gains of the
IFPL and FPL algorithms with parameters defined in Theorem~\ref{lemma-1}
satisfy the following inequalities
\begin{eqnarray}
l_{1:T}\ge e^{-\frac{2}{\mu}}r_{1:T}
\label{thh-1a}
\end{eqnarray}
for all $T$.

If ~${\rm Dev}(G)\le\frac{1}{2}\mu\delta$ for some $0<\delta<1$ then
\begin{eqnarray*}
l_{1:T}\ge (1-\delta)r_{1:T}
\end{eqnarray*}
holds for all sufficiently large $T$.
\end{corollary}
The second bound also holds for unbounded one-step gain games and so,
it is some improvement of results of~\cite{KaV2003} and~\cite{HuP2004}.

The following theorem, which is an analogue of the result from~\cite{KaV2003},
gives a bound for the IFPL algorithm
\begin{theorem}\label{IFPL-1}
The expected cumulative gain of the IFPL algorithm with the learning rate
(\ref{eps-2}) is bounded by
\begin{eqnarray}
r_{1:T}\ge\max\limits_{i=1,2} s^i_{1:T}-\frac{1}{\epsilon_T}
\label{ii-ff}
\end{eqnarray}
for all $T$.
\end{theorem}
The proof is along the line of the proof from~\cite{HuP2004} (which is
a refinement of the proof from~\cite{KaV2003}).

Let in this proof $s_t=(s^1_t,s^2_t)$ be a vector of one step gains and
$s_{1:t}=(s^1_{1:t},s^2_{1:t})$ be a vector of cumulative gains of two experts,
also let $\xi$ be a vector whose coordinates are random variables $\xi^1_t$
and $\xi^2_t$. Define $\epsilon_0=\infty$ and
$\tilde s_{1:t}=s_{1:t}+\frac{1}{\epsilon_t}\xi_t$ for $t=1,2,\dots$.
Consider the one-step gains $\tilde s_t=s_t+\xi_t\left(\frac{1}{\epsilon_t}-
\frac{1}{\epsilon_{t-1}}\right)$ for the moment.
For any vector $s$ and a unit vector $d$ denote
\begin{eqnarray*}
M(s)={\rm argmax}_{d\in D}\{d\circ s\},
\end{eqnarray*}
where $D=\{(0,1)^T,(1,0)^T\}$ is the set of two unit vectors of
dimension 2 and $\circ$ is the inner product of two vectors.

We first show that
\begin{eqnarray}\label{infis-1}
\sum\limits_{t=1}^T M(\tilde s_{1:t})\circ\tilde s_t\ge
M(\tilde s_{1:T})\circ\tilde s_{1:T}.
\end{eqnarray}
For $T=1$ this is obvious. For the induction step from $T-1$ to $T$
we need to show that
$$
M(\tilde s_{1:T})\circ\tilde s_T\ge M(\tilde s_{1:T})\circ\tilde s_{1:T}-
M(\tilde s_{1:T-1})\circ \tilde s_{1:T-1}.
$$
This follows from $\tilde s_{1:T}=\tilde s_{1:T-1}+\tilde s_T$ and
$M(\tilde s_{1:T})\circ\tilde s_{1:T-1}\le
M(\tilde s_{1:T-1})\circ \tilde s_{1:T-1}$.

We rewrite (\ref{infis-1}) as follows
\begin{eqnarray}\label{infis-2}
\sum\limits_{t=1}^T M(\tilde s_{1:t})\circ s_t\ge
M(\tilde s_{1:T})\circ\tilde s_{1:T}-
\sum\limits_{t=1}^T M(\tilde s_{1:t})\circ\xi_t\left(\frac{1}{\epsilon_t}-
\frac{1}{\epsilon_{t-1}}\right).
\end{eqnarray}
By the definition of $M$ we have
\begin{eqnarray}
M(\tilde s_{1:T})\circ\tilde s_{1:T}\ge M\left(s_{1:T}+
\frac{1}{\epsilon_T}\right)
\circ\left(s_{1:T}+\frac{\xi}{\epsilon_T}\right)=
\nonumber
\\
\max_{d} \{d\circ s_{1:T}\}+M\left(s_{1:T}+\frac{1}{\epsilon_T}\right)
\circ\frac{\xi_T}{\epsilon_T}.
\label{term-1}
\end{eqnarray}
The expectation of the last term in (\ref{term-1}) is equal
to $\frac{1}{\epsilon_T}$.
We have also
\begin{eqnarray}
\sum\limits_{t=1}^T \left(\frac{1}{\epsilon_t}-
\frac{1}{\epsilon_{t-1}}\right)M(\tilde s_{1:t})\circ\xi_t\le
\nonumber
\\
\sum\limits_{t=1}^T \left(\frac{1}{\epsilon_t}-
\frac{1}{\epsilon_{t-1}}\right)M(\xi_t)\circ\xi_t.
\label{w-1}
\end{eqnarray}
We have $P\{\max\xi>y\}\le P\{\xi^1>y\}+P\{\xi^2>y\}=2e^{-y}$.
Since $$
E(M(\xi_t)\circ\xi_t)=E(\max\{\xi^1,\xi^2\})\le\int_0^\infty 2e^{-y}dy=2,
$$
the expectation of (\ref{w-1}) has upper bound $\frac{2}{\epsilon_T}$.
Combining the bounds (\ref{infis-2})-(\ref{w-1}) we obtain (\ref{ii-ff}).
$\triangle$.

\begin{corollary}\label{cor-1}
Let $\mu$, $0<\mu<1$, be given. If the game of two experts is non-degenerative
then the expected cumulative gain of the IFPL algorithm is bounded by
\begin{eqnarray*}
r_{1:T}\ge\max\limits_{i=1,2}s^i_{1:T}(1-\mu).
\end{eqnarray*}
\end{corollary}

\section{Zero sum games}\label{negative-loss}

We consider a simplest example of the game of prediction with expert advice
with arbitrary positive and negative one-step gains and losses.
We apply these results in Appendix~\ref{brown-1}.

We consider a game $G$ of two experts with zero sum, i.e., $s^1_t=-s^2_t$
at each step $t$ of the game. If a one-step gain is negative it is
called a loss.
There are no restrictions on the absolute values of $s^1_t$.
Define {\it a volume} of the game at step $t$
$$
V_t=\sum\limits_{j=1}^t|s^1_j|.
$$
A game with zero sum is called {\it non-degenerate} if
$\lim\limits_{t\to\infty}V_t=\infty$.
Analogously to (\ref{dev-1}) we consider the deviation of the game $G$ with
zero sum
\begin{eqnarray*}\label{dev-2}
{\rm Dev}(G)=\limsup\limits_{t\to\infty}\frac{s_t}{V_t},
\end{eqnarray*}
where $s_t=|s^1_t|$ and $V_t$ is the volume of the game.

Evidently, the expected cumulative gain of the algorithm which
chooses one of two experts with probability $\frac{1}{2}$ equals
zero.

The following proposition is an analogue of Proposition~\ref{prop-1}.
\begin{proposition}\label{prop-2}
For any probabilistic algorithm of following the best expert,
two experts exist such that
the expected cumulative gain of this algorithm $E(s_{1:t})\le 0$
and two experts exist such that $E(s_{1:t})\ge 0$ for all $t$.
\end{proposition}
{\it Proof}. If $P\{I=1\}>\frac{1}{2}$ define $s^1_t=1$, $s^2_t=-1$
and define $s^1_t=-1$, $s^2_t=0$ otherwise. The following estimates are
analogous to that given in the proof of Proposition~\ref{prop-1}.
$\triangle$

The following theorem which is an analogue of the Theorem~\ref{perf-1}
for games with zero sum shows that if a probabilistic algorithm of the
following the leader has high performance in games with bounded one-step
gains then its expected cumulative gain in some games with unbounded
one-step expert gains can be arbitrary negative.

\begin{theorem}\label{perf-1a}
Let $L_t$ be any sequence of positive real numbers, $t=1,2,\dots$.
Let $\delta,\delta'$ be arbitrary close and arbitrary small
positive real numbers such that $\delta>\delta'$, and
let for any two experts with bounded one-step gains $s_t$,
i.e. such that $0\le s_t\le 1$ for all $t$, the expected cumulative gain
of the master algorithm has the lower bound
\begin{eqnarray}\label{perf-2f}
E(s_{1:t})\ge (1-\delta)|s^1_{1:t}|
\end{eqnarray}
for all sufficiently large $t$.
Then there exist two non-degenerate experts with unbounded one-step
gains such that
the expected performance of the master algorithm is bounded from above
\begin{eqnarray}\label{perf-3a}
E(s_{1:t})\le 2\delta'|s^1_{1:t}|-(1-2\delta')V_t
\end{eqnarray}
and such that $V_t\ge L_t$ for infinitely many $t$,
where $V_t$ is the volume of the game.
\end{theorem}

{\it Proof}. The proof is similar to the proof of
Theorem~\ref{perf-1}. It uses a modified version of
Lemma~\ref{l-1} which is also valid for negative gains with some
evident modifications. \footnote { A modified version of
Lemma~\ref{l-1} looks as follows: Let $\delta,\delta'$ be positive
real numbers such that $\delta'>\delta$, and let for any two
experts with bounded one-step gains (\ref{perf-2f}) holds for all
sufficiently large $t$. Then for any number $\tilde s^1$ a number
$s^1>0$ exists such that $s^1\ge\tilde s^1$ and $P\{I=1\}\ge
1-\delta'$, where $P\{I=1\}=f(s^1,-s^1)$. }

Let a master algorithm be given.
We define two experts with unbounded one-step gains as follows.
Define $s^1_1=s^2_1=0$. By modified version of Lemma~\ref{l-1} a number
$s^1$ exists such that $s^1>0$ and $P\{I=1\}\ge 1-\delta$, where
$P\{I=1\}=f(s^1,-s^1)$.

Let $t$ be even, and let $s^1_{1:t-1}$ and $s^2_{1:t-1}=-s^1_{1:t-1}$
be defined on previous steps. We will use the induction hypothesis:
$s^1_{1:t-1}>0$ and
$P\{I=1\}\ge 1-\delta$, where $P\{I=1\}=f(s^1_{1:t-1},s^2_{1:t-1})$.

Define one-step gains of experts 1 and 2: $s^1_t=-M_t$ and
$s^2_t=M_t$, where
$$
M_t=\max\left\{\frac{|E(s_{1:t-1})|}{2(\delta-\delta')},
L_t,\frac{V_{t-1}}{\delta}\right\}.
$$

Let $t$ be odd. By modified version of Lemma~\ref{l-1} a number
$s^1$ exists such that $s^1>|s^1_{1:t-1}|$ and
$P\{I=1\}\ge 1-\delta$, where $P\{I=1\}=f(s^1,-s^1)$.
Define
$s^1_t=s^1-s^1_{1:t-1}$, then $s^1_{1:t}=s^1$, and
$s^2_t=-s^1_t$. Evidently, the
induction hypothesis is valid after odd step $t$.

Let us prove that this construction is correct.
Let $t$ be even. Then by the induction hypothesis $s^1_{1:t-1}>0$
and $P\{I=1\}\ge 1-\delta$ (and $P\{I=2\}<\delta$).
By definition $s^1_{1:t}=s^1_{1:t-1}-M_t$ and $s^2_{1:t}=s^2_{1:t-1}+M_t$.
The expected one-step gain of the master algorithm is bounded
\begin{eqnarray*}
E(s_t)=s^1_t P\{I=1\}+s^2_t P\{I=2\}
\le -(1-\delta) M_t+\delta M_t=-(1-2\delta)M_t.
\end{eqnarray*}
By definition $L_t\le M_t\le V_t=V_{t-1}+M_t\le (1+\delta)M_t$. Then
\begin{eqnarray*}
E(s_{1:t})\le E(s_{1:t-1})-(1-2\delta) M_t\le -(1-2\delta')M_t=
\\
-M_t+2\delta'M_t\le -(1-\delta)V_t+2\delta' |s^1_{1:t}|
\end{eqnarray*}
for all even steps $t$.
$\triangle$

We consider the {\it non-degenerate games}, i.e.,
such that $V_t$ is unbounded.

To obtain the lower bounds we reduce our zero sum game to a game
with non-negative one-step gains.
Define one-step gain of new experts $\tilde s^i_t=s^i_t+|s^1_t|$ for $i=1,2$.
Then $\tilde s^i_t\ge 0$ for all $t$ and $\tilde s^1_t=0$ or
$\tilde s^2_t=0$ for all $t$. By definition $\tilde s^i_{1:t}=s^i_{1:t}+V_t$
for $i=1,2$, where $V_t$ is the volume of the initial game.
Evidently, the FPL and IFPL algorithms defined in Section~\ref{sect-3}
make the same choices for experts of both type.

The expected one-step gains of the master algorithm
for for experts of both type satisfy
$\tilde l_t=s^1_tP\{I=1\}+s^2_tP\{I=1\}+|s_t|$. This implies
the equality $\tilde l_{1:t}=l_{1:t}+V_t$ for expected cumulative gains.
The analogous equalities hold for
$\tilde r_t$, $\tilde r_{1:t}$ and $r_t$, $r_{1:t}$.

The following theorem is a corollary of Theorem~\ref{cor-2}.

\begin{theorem}\label{cor-2f}
For any $\mu$ such that $0<\mu<1$ an FPL algorithm can be
specified such that for any non-degenerate game of two experts
its expected cumulative gain at any step $T$ has the lower bound
\begin{eqnarray}
l_{1:T}\ge e^{-\frac{2}{\mu}}(1-\mu)|s^1_{1:T}|-
V_T(1-e^{-\frac{2}{\mu}}(1-\mu)),
\label{zero-1}
\end{eqnarray}
where $s^1_{1:T}$ is the cumulative gain of the the first expert and $V_T$
is the volume of the game at step $T$.

If ~${\rm Dev}(G)\le\frac{1}{2}\mu\delta$ for some $0<\delta<1$ then
\begin{eqnarray}
l_{1:T}\ge (1-\delta)(1-\mu)|s^1_{1:T}|-(\delta+\mu) V_T
\label{thh-2f}
\end{eqnarray}
holds for all sufficiently large $T$.
\end{theorem}
{\it Proof}.
This theorem follows from Theorem~\ref{cor-2}
and relations between one-step gains $\tilde s^i_t$ and $s^i_t$,
$i=1,2$, of two type of experts.
$\triangle$

{\bf Remark}.
In case when ~${\rm Dev}(G)\le\frac{1}{4}\mu\delta$,
the bound (\ref{thh-2f}) can be improved for some $t$ if we
replace in Section~\ref{sect-3} the learning rate (\ref{eps-1}) on
$$
\epsilon_{t-1}=\frac{1}{\mu\max\limits_{j<t}|s^1_{1:j}|}.
$$
Then the inequality (\ref{thh-2f}) can be obtained directly (without using
the modified experts) from inequalities (\ref{i-00}) and (\ref{i=0}).
We can prove that for any $T$ such that
$T=arg\max\limits_{t\le T}|s^1_{1:t}|$
\begin{eqnarray*}
l_{1:T}\ge (1-\delta)(1-\mu)|s^1_{1:T}|.
\end{eqnarray*}

\appendix

\section{Learning the fractional Brownian motion}\label{brown-1}
In this section we present some example of the zero sum game
studied in Section~\ref{negative-loss}. Rogers~\cite{Rog}, Delbaen
and Schachermayer~\cite{DeS}, and Cheredito~\cite{che} have
constructed arbitrage strategies for a financial market that
consists of money market account and a stock whose price follows a
fractional Brownian motion (for continuous time) with drift or an
exponential fractional Brownian motion with drift. Vovk~\cite{Vov}
has reformulated these strategies for discrete time.

Let $S_0,S_1,\dots , S_n,\dots$ be a sequences of prices of
some financial instruments such as stocks or bonds. We consider the
following ``financial'' game between an investor and the
market. The investor can use the long and short selling.

\noindent FOR $t=1,2,\dots T-1$

At the beginning of trading period
the investor's cumulative income (or loss) earned from the beginning
of the game is $s_{1:t-1}=\sum_{i=1}^{t-1} s_i$.

At the beginning of trading period, observing his past incomes and losses
the investor determines
the number $C_t$ of shares of the stock needed to realize his strategy.

At the end of trading period the market discloses the price
$S_{t+1}$ of the stock, and the investor incur his current income
or loss at the period $t$ \footnote { We suppose that this price
is also valid at the beginning of the period $t+1$. }
$$
s_t=C_t(S_{t+1}-S_t).
$$
\noindent ENDFOR

Denote $\Delta S_t=S_{t+1}-S_t$. We have the following equality
\begin{eqnarray}\label{main-1}
(S_T-S_0)^2=(\sum\limits_{t=0}^{T-1}\Delta S_t)^2=\sum\limits_{t=0}^{T-1}
2(S_t-S_0)\Delta S_t+\sum\limits_{t=0}^{T-1}(\Delta S_t)^2.
\end{eqnarray}

The equality (\ref{main-1}) leads to the two strategies which are represented
by two experts: At the beginning of step $t$
Expert 1 holds the number of shares
\begin{eqnarray}
C^1_t=2C(S_t-S_0),
\label{stat-1}
\end{eqnarray}
Expert 2 holds the number of shares
\begin{eqnarray}
C^2_t=-2C(S_t-S_0),
\label{stat-2}
\end{eqnarray}
where $C$ is an arbitrary positive constant.

These strategies at step $t$ earn the incomes
$s^1_t=2C(S_t-S_0)\Delta S_t$ and
$s^2_t=-s^1_t$.
The strategy (\ref{stat-1}) earns in $T$ steps of the game the income
$
s^1_{1:T}=2C((S_T-S_0)^2-\sum\limits_{t=1}^{T-1}(\Delta S_t)^2).
$
The strategy (\ref{stat-2}) earns in $T$ steps the income
$s^2_{1:T}=-s^1_{1:T}$.

The number of shares $C^1_t=2C(S_{t-1}-S_0)$ in the strategy (\ref{stat-1})
or number of shares $C^2_t=-2C(S_{t-1}-S_0)$ in the strategy (\ref{stat-2})
can be positive or negative. Expert 1 uses the hypothesis that
the Hurst exponent of the price of stock is $>\frac{1}{2}$
(a smoother trend). Expert 2 uses the hypothesis that
the Hurst exponent is $<\frac{1}{2}$ (volatility is high).

It is reasonable to derandomize the FPL algorithm for this
financial game. For that, the investor must follow both experts
strategies simultaneously holding $P\{I=1\}C^1_t+P\{I=2\}C^2_t$
shares of a stock at any step $t$. In this case
Theorem~\ref{cor-2f} holds, where the expected gain at step $t$
is replaced on a pure gain $s_t=P\{I=1\}C^1_t\Delta
S_t+P\{I=2\}C^2_t\Delta S_t$. \footnote { Analogously we can
derandomize all probabilistic games of this paper if we allow
for Learner to receive a given fraction of the gain of an
expert. }


\begin{thebibliography}{99}

\bibitem{AAAGO2006}
Chamy Allenberg, Peter Auer, Laszlo Gyorfi and Gyorgy Ottucsak:
Hannan Consistency in On-Line Learning in Case of Unbounded Losses Under Partial Monitoring.
LNCS, Volume 4264, 229-243, Springer-Verlag Berlin Heidelberg 2006.

\bibitem{DeS}
Delbaen F., Schachermayer W.: A general version of the fundamental
theorem of asset pricing. Mathematische Annalen, 300 (1994), 463-520.

\bibitem{che}
Cheredito P.: Arbitrage in fractional Brownian motion, Finance and Statistics,
7 (4) (2003), 533-553.

\bibitem{Han57}
Hannan J.: Approximation to Bayes risk in repeated plays. In M. Dresher,
A.W. Tucker, and P. Wolfe, editors, Contributions to the Theory of Games 3,
97-139, Princeton University Press, 1957.

\bibitem{HuP2004}
Hutter M., Poland J.: Prediction with expert advice
by following the perturbed leader
for general weights,
(S.Ben-Dawid, J.Case, A.Maruoka (Eds.)): ALT 2004
LNAI 3244, 279-293. Springer-Verlag Berlin Heidelberg 2004.

\bibitem{PoHP2005}
Poland J., Hutter M.: Defensive universal learning with experts.
for general weight.
(S.Jain, H.U.Simon and E.Tomita (Eds.)): ALT 2005
(S.Jain, H.U.Simon and E.Tomita (Eds.)),
LNAI 3734, 356-370. Springer-Verlag Berlin Heidelberg 2005.

\bibitem{KaV2003}
Kalai A., Vempala S.: Efficient algorithms for online decisions.
In Proceedings of the 16th Annual Conference on Learning Theory (COLT-2003),
LNAI, 506-521, Berlin, 2003, Springer.
Extended version in Journal of Computer and System Sciences, 71, 2005, 291-307.

\bibitem{Rog}
Rogers C.: Arbitrage with fractional Brownian motion. Mathematical Finance,
7 (1997), 95-105.

\bibitem{Vov}
Vovk V.: A game-theoretic explanation of the $\sqrt{dt}$ effect,
Working paper $\#5$, 2003, http://www.probabilityandfinance.com


\end{thebibliography}
\end{document}